\documentclass[letterpaper, 10 pt, conference]{ieeeconf}  

\IEEEoverridecommandlockouts                              

\usepackage{cite}
\usepackage{amsmath,amssymb,amsfonts}
\usepackage{algorithmic}
\usepackage{graphicx}
\usepackage{textcomp}
\usepackage{xcolor}
\usepackage{fancyhdr}

\title{\LARGE \bf
Autonomous Planning In-space Assembly Reinforcement-learning free-flYer (APIARY) International Space Station Astrobee Testing
}

\author{Samantha Chapin$^{1*}$, Kenneth Stewart$^{1*}$, Roxana Leontie$^{1}$, and Carl Glen Henshaw$^{1}$
\thanks{*These authors contributed equally to this work}
\thanks{$^{1}$ Authors are with the U.S. Naval Research Laboratory Naval Center for Space Technology, Washington, DC, United States
        {\tt\small samantha.h.chapin.civ@us.navy.mil}}
}

\begin{document}

\maketitle

\thispagestyle{empty}
\pagestyle{empty}

\begin{abstract}
The US Naval Research Laboratory's (NRL's) Autonomous Planning In-space Assembly Reinforcement-learning free-flYer (APIARY) experiment pioneers the use of reinforcement learning (RL) for control of free-flying robots in the zero-gravity (zero-G) environment of space. 
On Tuesday, May 27th 2025 the APIARY team conducted the first ever, to our knowledge, RL control of a free-flyer in space using the NASA Astrobee robot on-board the International Space Station (ISS). 
A robust 6-degrees of freedom (DOF) control policy was trained using an actor-critic Proximal Policy Optimization (PPO) network within the NVIDIA Isaac Lab simulation environment, randomizing over goal poses and mass distributions to enhance robustness.  This paper details the simulation testing, ground testing, and flight validation of this experiment. This on-orbit demonstration validates the transformative potential of RL for improving robotic autonomy, enabling rapid development and deployment (in minutes to hours) of tailored behaviors for space exploration, logistics, and real-time mission needs.
\end{abstract}


\section{Introduction}
Robotic spacecraft play a crucial role in scientific exploration. 
Robotic rovers have operated on Mars continuously for two decades and have revolutionized our understanding of the climatic and geological history of Earth's neighbor. 
The Hubble Space Telescope, the James Webb Space Telescope, and many other orbital observatories have produced an enormous amount of astronomical data and have had a profound impact on our understanding of the universe. 
More recently, robotic spacecraft such as the SpaceLogistics Mission Extension Vehicle (MEV) have demonstrated the ability of space robots to autonomously dock to and extend the life of valuable communications satellites~\cite{pyrak2022performance}, and the upcoming DARPA Robotic Servicing of Geosynchronous Satellites (RSGS) spacecraft will be the world's first orbiting satellite mechanic, being able to inspect, relocate, and upgrade satellites in geosynchronous orbit using its two 7-DOF robotic arms and interchangeable end-effectors~\cite{duke2021orbit}~\cite{roesler2017orbital}.

Despite these advancements, current robotic spacecraft require significant, and often expensive, human ground control. Spacecraft autonomy lags the state-of-the-art significantly. 
RSGS, for instance, uses Resolved Rate Motion Control (RRMC) inverse kinematics that requires human generated 6-DOF waypoints to perform task space operations~\cite{henshaw2014darpa}. 
Generating and verifying a single RSGS task space operation takes hours to days of a highly qualified robotics operator's time. 
This is because spacecraft computers are significantly slower than those used in other domains - the RSGS flight computer consists of five 800 MHz SBCs - and because ``classic'' robotic automation techniques such as linear control, inverse kinematics, waypoint-based path planning, and so on can be mathematically analyzed for correctness. 
This is a vitally important feature of such algorithms given that spacecraft are among the most expensive engineered objects in existence; even a relatively mundane geosynchronous communications satellites costs hundreds of millions of dollars, and the James Webb Space Telescope cost more than ten billion dollars. 
It is therefore crucial that the flight computers controlling these spacecrafts be highly reliable and that their software is as stable and as well understood as possible.

Currently, the application of RL to spacecraft control presents challenges in ensuring the same level of formal verification as traditional methods. This work focuses on demonstrating the feasibility of RL for free-flyer control, but future research will be necessary to address the critical need for verifiable safety guarantees.

However, the next generation of robotic spacecraft will require significantly more autonomy. This increased capability is essential for enabling new exploration missions, like NASA's SWIM mission to explore Europa's ocean with submersibles. Greater autonomy also reduces the cost and increases the speed of orbital operations, such as assembling large space telescopes or solar power beaming stations.
In this vein, RL approaches to the control of free-flying spacecraft, especially as a precursor to RL-based control of free-flying spacecraft with robotic arms, have great potential to revolutionize the capability of space robotic systems.
This work demonstrates an RL-based 6-DOF control policy for a free-flying robotic spacecraft: NASA's Astrobee robot operating on-board the ISS. 
These cube-shaped robots, measuring 12.5 inches (30.5 cm) on a side, utilize electric fan propulsion to navigate in 6-DOF, use cameras to perceive their environment, and have a 2-DOF arm for perching, inspection, and servicing.  

We've bridged the sim-to-real (Sim2Real) gap for the Astrobee platform by using the NVIDIA Isaac Lab simulation environment. The Sim2Real gap, arises from discrepancies between the simulated and real-world dynamics, often leading to degraded performance and necessitates extensive real-world fine-tuning. While RL offers the potential for adaptable and optimized control policies, overcoming this Sim2Real gap is critical for realizing its benefits in space-based applications. Within Isaac Lab, a robust and capable Divert-and-Attitude Control System (DACS) is trained for Astrobee under ISS-like conditions, including zero-G and then verified within a simulation of the ISS interior. 
Robustness is enhanced by randomizing over goal position, goal orientation, and in some experiments, mass. 
The policy generates wrenches that are executed by Astrobee's onboard fan-based thrusters and existing wrench control logic. 
Simulations, along with tests on the Astrobee ground air bearing simulator hardware, demonstrate the resulting RL policy's ability to execute motion trajectories, culminating in the (to our knowledge) first RL control of a free-flyer in space using the NASA Astrobee robot on-board the ISS.

\subsection{Related Work}

Traditional spacecraft control relies heavily on classical control methods, such as Proportional-Derivative (PD) and Proportional-Integral-Derivative (PID) controllers for attitude determination, and bang-bang control for divert maneuvers~\cite{grappling_spacecraft}. 
These methods have proven effective for basic stabilization and trajectory adjustments. 
However, the increasing complexity of space missions, including intricate robotic arm operations and autonomous maneuvering, require the exploration of more advanced control strategies. 
Recently, RL has emerged as a promising approach for enhancing spacecraft control, offering the potential for greater autonomy, adaptability, and optimized performance in challenging space environments~\cite{vedant2019rlsac}. 

Most modern spacecraft are three-axis stabilized, meaning they have the ability to maintain a desired orientation with respect to the earth and the sun (or, for space telescope, with respect to an astronomical object being observed). 
This is necessary to maintain the solar incidence angle of the solar panels, high bandwidth communications links with satellite dishes on the earth, and fields of view for the satellite's science or imagery sensors, if any~\cite{larson1999space}. 
Attitude control is typically accomplished with standard PD or PID linear control~\cite{wertz2012spacecraft}. 
Typical spacecraft rotational actuators are reaction wheels, rate-moment gyros, or in some cases magnetic torquer bars that react against the earth's magnetic field~\cite{Wertz_Larson_2010}\cite{wertz2012spacecraft}. 
These are proportional output devices, and standard linear control theory applies well when using them. 
Spacecraft typically have significant flexible body dynamics due to large solar panels and, sometimes, to large radio antennas~\cite{wie1998space}. 
They often have significant dynamics due to fuel slosh, the fuel used for their thrusters exerting nontrivial dynamics in ways that are nonintuitive due to the microgravity environment~\cite{wie1998space}. 
These dynamics are typically treated as disturbances to the attitude control system, and hence robust linear control techniques are often employed. 
This has the advantage of resulting in attitude control software that can be mathematically analyzed for correctness and for performance, at the cost of (often significant) reductions in performance. 
It is not unusual for a large spacecraft slew to take tens of minutes to hours. 
``Reaction control'', the ability of the spacecraft to maintain its translational position, e.g. to stay in an orbit, to modify its orbit, or to approach or avoid another spacecraft, relies on thrusters, which have a minimum impulse time, and are hence inherently nonlinear actuators. 
Bang-bang control is typically used for reaction control~\cite{philip2003relative}. 
Attitude and reaction control are usually considered separately. 
While effective for established mission profiles, these classical approaches often require extensive human intervention and may struggle with dynamic complexities of advanced maneuvers and robotic interactions.

Major spacecraft motions are usually planned by human controllers; this includes orbit modifications and large ``slews'', significant rotations of a spacecraft to e.g. observe a different part of the sky or point to different spots on the earth. 
Automated decision making for slewing a spacecraft appears to have been first implemented on NASA's Deep Space-1 (DS-1) spacecraft, which was in large part a demonstrator for spacecraft automation. 
DS-1 used an onboard constraint satisfaction planner to schedule different science observation objectives and to plan slews to satisfy these objectives. DS-1 relied on standard linear control laws to execute these maneuvers~\cite{rayman2000results}.

More advanced types of robotic control such as model-predictive control or RL do not appear to have been demonstrated on any real spacecraft, but there is related work attempting to advance this state-of-the-art~\cite{dlr189550}~\cite{sprecht2023rendevous}~\cite{WU2020105657}. Even well understood nonlinear robotic control techniques such as computed torque control or Hamiltonian passivity control have not so far been used for any known robotic spacecraft missions~\cite{sanner1995}~\cite{WANG201276}.
However, RL for satellite DACS and for the combined control of attitude/divert and robotic arms has been investigated in simulation~\cite{elkins2020autonomous}. 
Moving beyond simulations, this work focuses on applying RL to real world spacecraft platforms, demonstrating its potential for practical implementation.

\section{Reinforcement Learning Simulation Training to Experimental Validation Pipeline}
\label{sec:methods}

\begin{figure}[h]
    \centering 
    \includegraphics[width=\linewidth]{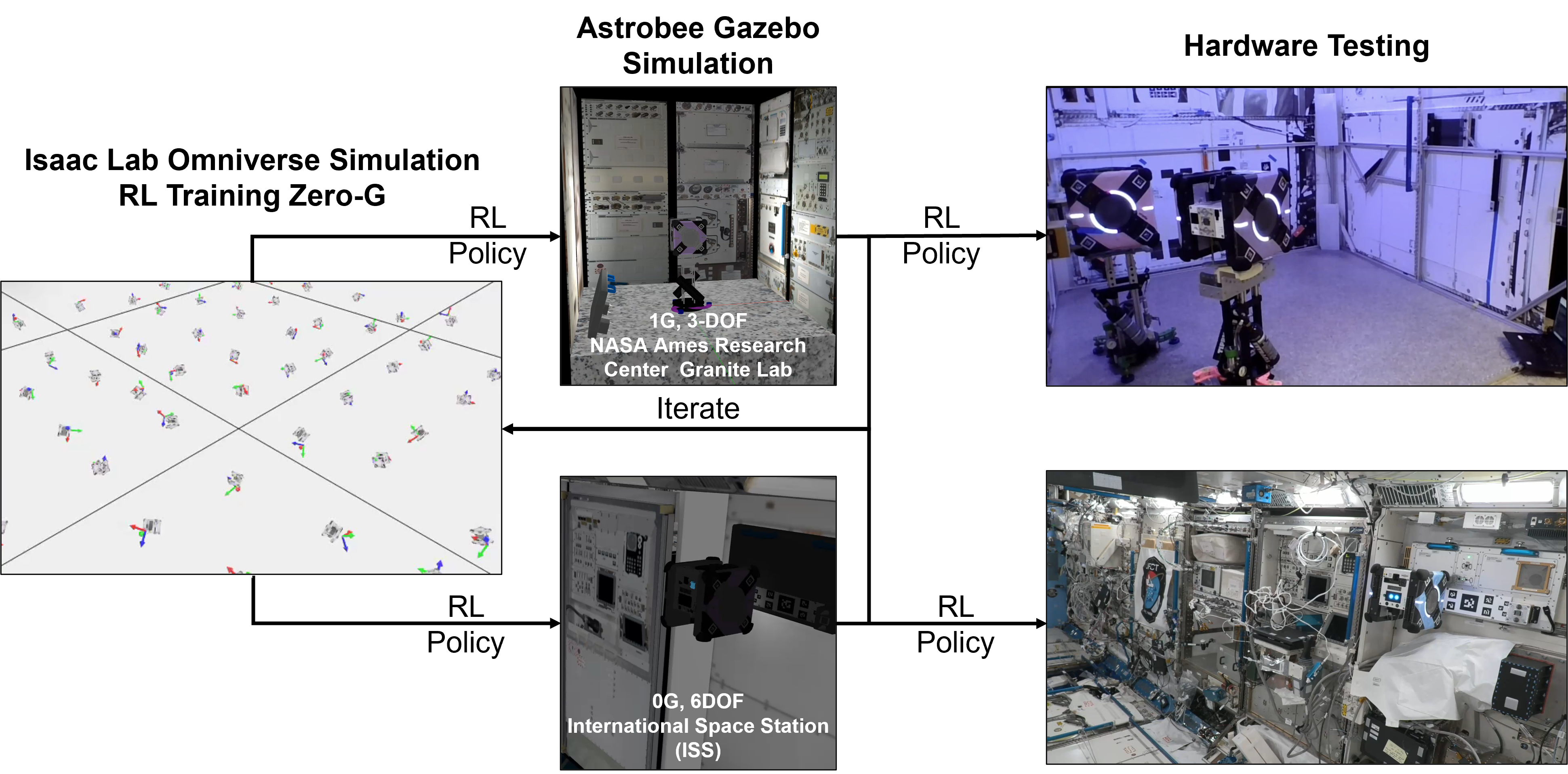}
    \caption{Reinforcement learning simulation to ground and flight testing flowchart. Train the Astrobee in zero-G environment, and use the same policy to test in simulation and hardware testing. Iterate by re-training a new policy if necessary based on simulation and hardware testing results.}
    \label{fig:pipeline} 
\end{figure}

An RL controller for the Astrobee free-flyer was trained in NVIDIA Omniverse with Isaac Lab and then tested in the NASA Astrobee Robot Software simulator in Gazebo and on hardware at the Granite Lab and on-board the ISS. 
By mimicking the Astrobee's ground and flight testing environments, this work seeks to train RL policies that achieve optimal robot control that is robust to environment changes. 
Figure \ref{fig:pipeline} illustrates the workflow for generating an RL policy through simulation in a zero-G environment and testing across various environments. 
These RL policies were loaded into the Astrobee simulation environment and used to control the robot's motion in simulation and for ground testing. This initial development stage verifies this RL policy can safely control the Astrobee movement before testing on the ISS. The simulators and testing facilities used for training and testing are described in more detail below. 

\subsection{Training: NVIDIA Omniverse}
\begin{figure}[h]
    \centering
    \includegraphics[width=0.9\linewidth]{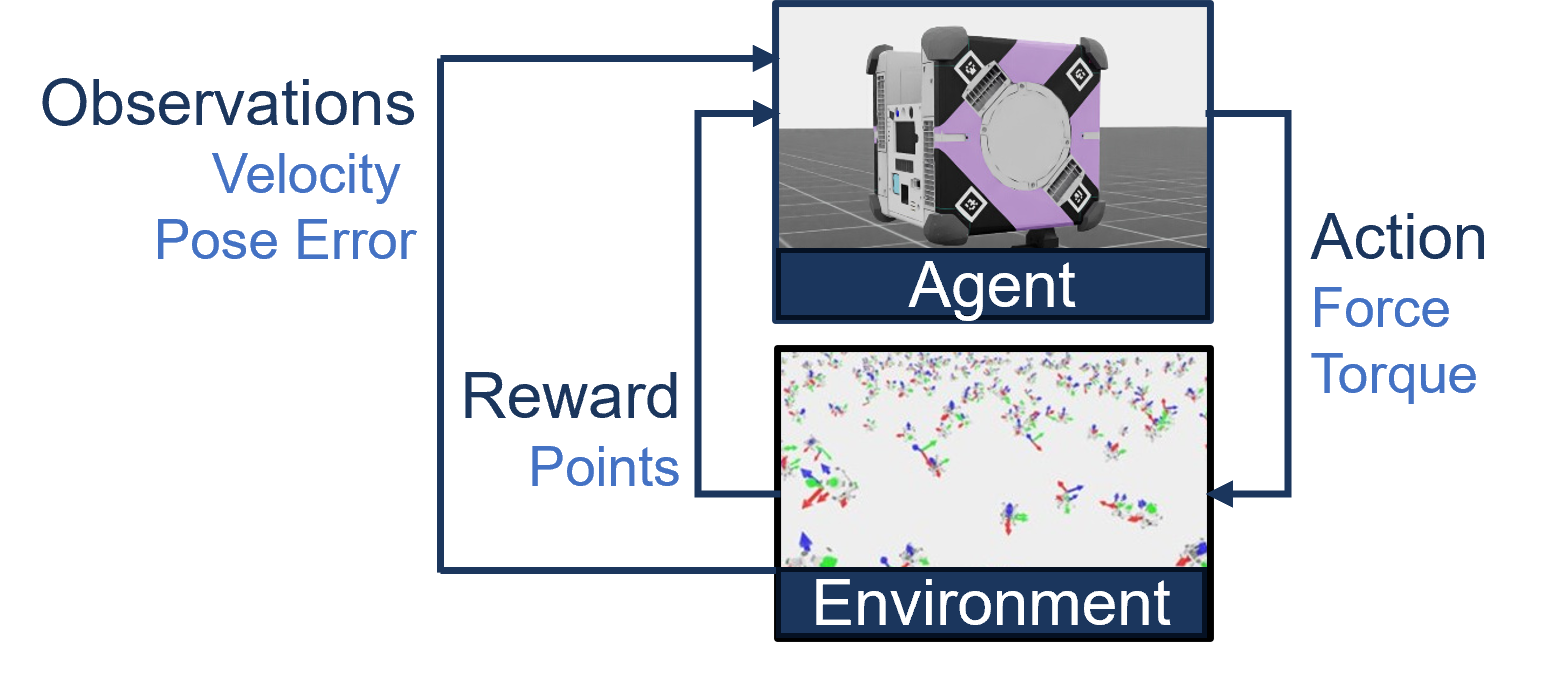}
    \caption{Diagram showing RL training process with Astrobee learning to reach a desired end pose based on observations (the robot's linear and angular velocity and positional and orientation error) and rewards (points for reducing pose error and minimizing velocity), commanding actions (force and torque) resulting in Astrobee motion.}
    \label{fig:RL_diagram}
\end{figure}

The NVIDIA Omniverse physics simulator allows for altering and randomizing various physics aspects of the environment and agents within the simulation, e.g., mass, friction, and gravity. 
This variation can change how agents behave and interact with the environment~\cite{naveed2024omniverse_survey}\cite{mittal2023orbit}. 
Figure \ref{fig:RL_diagram} shows the training environment used in Omniverse: a zero-G environment where Astrobee is a free-flyer with no other objects in the environment, since the policy does not rely on vision for control. A model of the Astrobee~\cite{astrobee_model} was exported from the Astrobee Gazebo simulation~\cite{Nasa}, including physics properties such as mass, and imported into IsaacLab for training. 
During training, we exploit the Omniverse setup for utilizing GPUs to run many parallel robotic simulations to speed up training and increase robustness by varying parameters within each environment such as the mass of a robotic agent to account for unexpected variation in real world conditions~\cite{rudin2022learning}. 
The RL policy was trained and tested with the following general setup: 

\begin{enumerate}
    \item \textbf{Goal}: Move to Desired End Pose
    \item \textbf{Observations}: Position Error, Orientation Error, Linear Velocity, Angular Velocity
    \item \textbf{Actions}: Force, Torque
    \item \textbf{Reward}: Positive Points for Reducing Pose Error, Negative Points for High Velocity Motion
\end{enumerate}

We train deep reinforcement learning (DRL) policies to learn to control the robot in these simulated environments and then test them in the real world on hardware. For more information on the algorithms and curriculums used to train the APIARY RL policies, see our associated paper~\cite{2025_APIARY_iSpaRo_Paper_2}. 

\subsection{Simulation Testing: NASA Astrobee Robot Software Simulator}
The NASA Astrobee Robot Software~\cite{Nasa} was developed by NASA Ames Research Center and allows researchers to test out algorithms in simulation before hardware testing at their Granite Lab or on the ISS, shown in Figure \ref{fig:sim_granite} (a) and Figure \ref{fig:sim_iss} (a) respectively. 
The NASA Astrobee simulation environment uses the Gazebo simulator. 
In this setup a simulated Astrobee can be commanded using pre-defined commands for localization, navigation, control, perching arm manipulation, etc. 
For testing the RL control policy, the standard Astrobee software was modified to accept force and torque commands directly from the RL policy, bypassing its internal control calculations. 
The inputs and outputs of the Guidance Navigation Control (GNC) standard control (CTL) function and the replacement of the CTL command for force and torque by the output action of the RL policy are shown below in Figure \ref{fig:replacing_astrobee_controller}.

\begin{figure*}[h]
    \centering
    \includegraphics[width=0.9\linewidth]{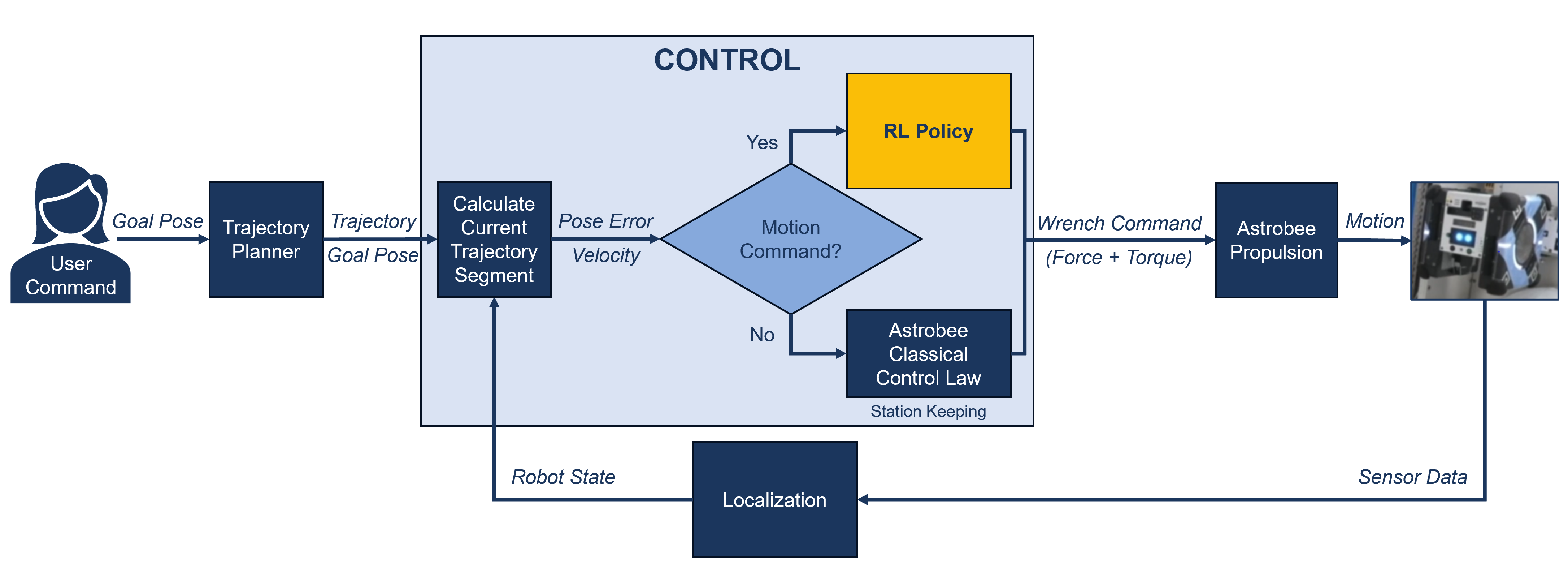}
    \caption{Diagram of the control architecture of the Astrobee motion, highlighting replacement of the baseline Astrobee controller with the RL policy, yellow box, when a movement is commanded.}
    \label{fig:replacing_astrobee_controller}
\end{figure*}

\subsection{Hardware Testing: NASA Ames Research Center Granite Lab and ISS}
The NASA Ames Research Center Granite Lab, shown in Figure \ref{fig:hardware_V2} (a) and (b), features a 2-meter square granite table testbed replicating an ISS environment (including ISS imagery, an Astrobee docking port, and an astronaut handrail)~\cite{astrobeeguest}. 
An Astrobee robot, mounted on air-bearings for frictionless motion across the granite surface, uses its on-board fan-based propulsion system for 3-DOF motion testing. 
NASA has three Astrobees on-board the ISS, for the final flight testing the Bumble Astrobee, blue version, was used autonomously without astronaut assistance, shown in Figure \ref{fig:flight} (a) and (b).
This allowed for zero-G, 6-DOF testing of the RL policy in a pressurized air environment. 
 
\section{Experimental Results for RL Policy Iteration}

This study evaluates the performance of a zero-G trained RL control policy for the Astrobee robot across various simulated and physical environments. 
The RL policy was trained in an NVIDIA Omniverse zero-G simulation and then tested before flight in four distinct environments: NVIDIA Omniverse zero-G simulation, Gazebo simulations of both the ISS and the NASA Ames Research Center Granite Lab, and the physical Granite Lab testbed. 
This diverse range of test environments allows for a comprehensive assessment of the RL policy's robustness and generalizability across varying levels of fidelity and real world conditions in preparation for the on-orbit validation aboard the ISS. 
Iteration of how the RL training was setup was completed throughout testing based on the performance in simulation and hardware experiments. 

The RL control of the Astrobee is compared against the baseline Astrobee controller~\cite{Astrobee2015}, developed and used for normal operations of the Astrobee by NASA Ames Research Center, and relies on the localization of the Astrobee~\cite{astrobee_localization}. 

\subsection{Simulation Testing: NASA Astrobee Robot Software Simulator}

Before testing on hardware the RL policy was verified on the NASA Ames Gazebo simulator and software suite~\cite{Nasa}. 
The baseline controller that outputs the force and torque commands was replaced with the RL policy, and the rest of the software was left unchanged. 
Before flight testing, NASA first verifies any Astrobee software modifications and experiments on the ground at the Granite Lab. 
The generated RL policy was verified both in the granite table simulation as well as the ISS simulation.

\subsubsection{One-G, 3-DOF, Granite Lab Simulation}

\begin{figure}[h]
    \centering 
    \includegraphics[width=\linewidth]{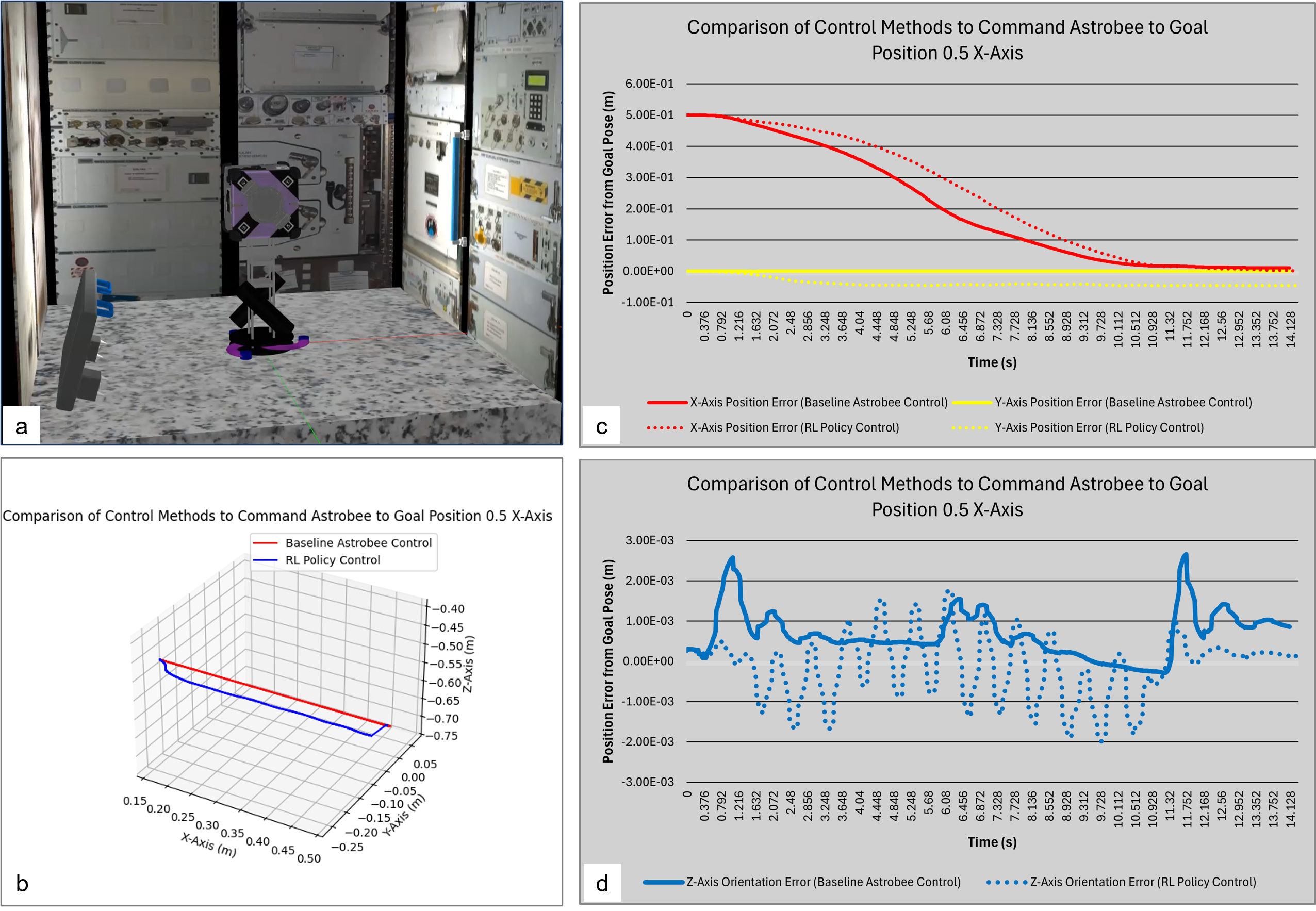}
    \caption{Granite Lab 1G Simulation Comparison (a) Astrobee simulator starting position. (b) 3D plot of baseline (red) and RL policy (blue) controllers' motion to undock Astrobee. (c) Position error from goal pose for Astrobee 0.5 X-Axis command using baseline Astrobee controller (solid lines) vs RL policy controller (dotted lines). (d) Orientation error from goal pose for Astrobee 0.5 X-Axis command.}
    \label{fig:sim_granite} 
\end{figure}

In the Granite Lab one-G simulation the Astrobee was commanded to move 0.5-meters in the X-axis direction. 
Figure \ref{fig:sim_granite} (a) shows the Astrobee in the starting pose at the center of the granite table and (b) shows the 3D plot of the motion of the Astrobee motion using the baseline NASA controller (red) and RL policy controller (blue). 
Figure \ref{fig:sim_granite} (c) and (d) shows the pose error, position and orientation respectively, respective to the goal pose over the time of the 0.5 X-Axis maneuver, comparing the Astrobee controller (solid lines) and the RL policy controller (dotted lines). 
Since this simulation is constrained to 3-DOF only the X and Y positions were plotted and only the Z orientation since those are the DOF that have free movement in this constrained setup.

The baseline Astrobee controller follows a straight path to complete the 0.5 X-axis maneuver while the RL policy controller has an additional Y-Axis error that is not present in the baseline. 
The controllers have slightly different trajectories for the X-Axis motion but converge on the correct X-Axis goal position, minimizing their final X-Axis position error. Both controllers experience variations in the Z-axis orientation throughout the maneuver. 

\subsubsection{Zero-G, 6-DOF, ISS Simulation}

\begin{figure}[h]
    \centering 
    \includegraphics[width=\linewidth]{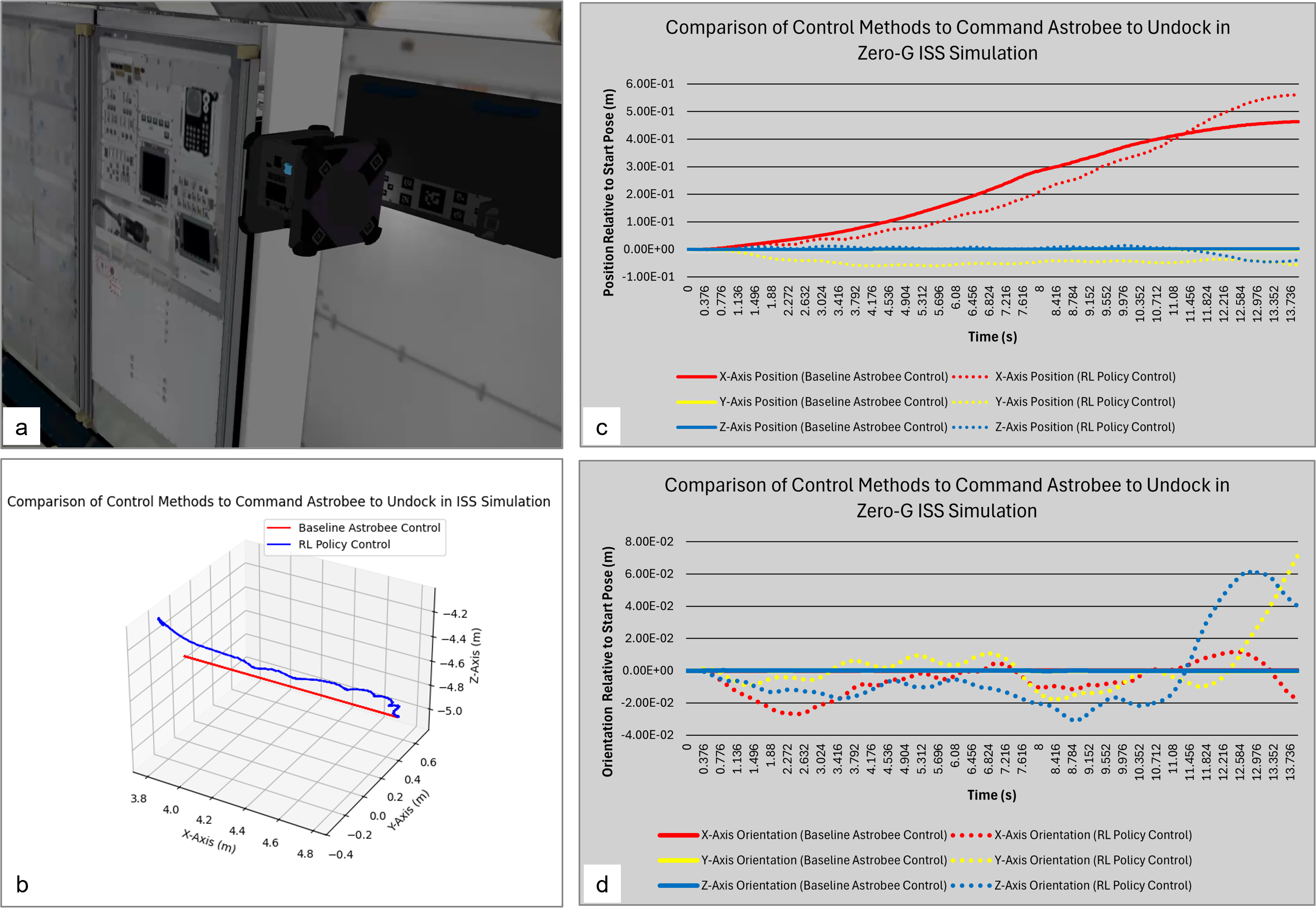}
    \caption{ISS Zero-G Simulation Comparison (a) Astrobee simulator starting position, docked. (b) 3D plot of baseline (red) and RL policy (blue) controllers' motion to undock Astrobee. (c) Position error from goal pose for Astrobee undock command using baseline Astrobee controller (solid lines) vs RL policy controller (dotted lines). (d) Orientation error from goal pose for Astrobee undock command.} 
    \label{fig:sim_iss} 
\end{figure}

In the ISS zero-G simulation the Astrobee was commanded to undock which results in a 0.5-meter displacement command in the X-axis direction. 
Figure \ref{fig:sim_iss} (a) shows the Astrobee in the docked starting pose and (b) shows the 3D plot of the motion of the Astrobee undocking using the baseline controlled (red) and RL policy controller (blue). 
Figure \ref{fig:sim_iss} (c) and (d) shows the pose error, position and orientation respectively, respective to the goal pose over the time of the undocking maneuver, comparing the Astrobee controller (solid lines) and the RL policy controller (dotted lines). 

The baseline controller ends the maneuver with less error compared to the goal position than the RL policy controller. The X-axis position error during the manuver is comparable between the two controllers, but the final error is minimized for the baseline controller while the RL policy slightly overshoots. 
Additionally, the Y and Z-axis errors are greater in the RL policy controller, and while the baseline controller has little orientation error the RL policy has significant orientation error. 

\subsection{Hardware Testing: NASA Ames Research Center Granite Lab}
\begin{figure}[h]
   \centering 
   \includegraphics[width=\linewidth]{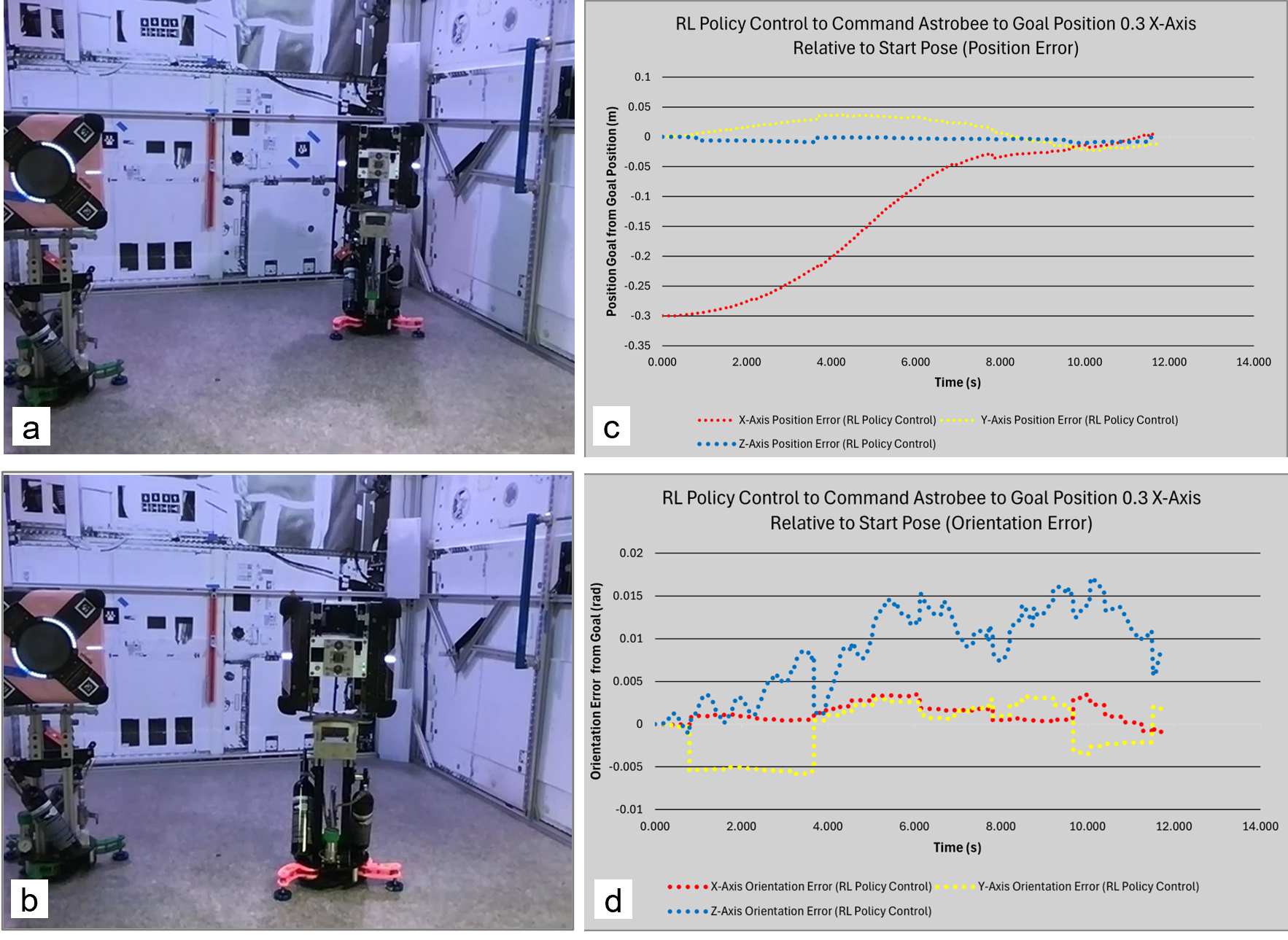}
   \caption{Granite Lab 1G Hardware Comparison (a) Astrobee starting position, docked. (b) Astrobee ending position, 0.3-meter displacement in the X-axis relative to starting position. (c) Position error from goal pose for Astrobee undock command using baseline Astrobee controller (solid lines) vs RL policy controller (dotted lines). (d) Orientation error from goal pose for Astrobee undock command.}
   \label{fig:hardware_V2}
\end{figure}

Initial testing was performed at the NASA Ames Granite Lab and verified that the RL policy could be loaded on-board the Astrobee hardware and execute commands to control the Astrobee's pose. 
Since the ground-test Astrobee is identical to its ISS counterpart, it was essential to verify that the RL algorithms could run within the flight hardware's computational limits. An example command to translate the Astrobee 0.3 m along its initial X-axis was executed, and the results are presented in Figure \ref{fig:hardware_V2}. 
In this experiment, the RL policy alone generated the motion commands, issuing wrench commands to the Force Allocation Module (FAM) that drives the fan-based propulsion system.
 
This test shows an example run where the RL policy successfully commanded the robot through a 0.3 m translation along its relative X-axis from the starting pose, completing the maneuver. 
Although it exhibited off-axis drift in Y and rotational deviations, consistent with simulation results, the resulting motion still fell within the Astrobee's defined success thresholds.  

\section{Flight Testing}

The APIARY experimentation culminated in testing on-board the ISS. 
approximately seven minutes were allocated for this testing, and within that time eight maneuvers were completed. These included an undock, rotations, translations, and two attempted redocks, with one successful redock. 
\begin{enumerate}
    \item Undock (+0.5 meter X-axis translation)
    \item Negative Rotation (-20 deg Z-axis)
    \item Positive Rotation (+20 deg Z-axis)
    \item Single-axis Translation (+0.5 X-axis)
    \item Pre-docking Motion to Dock Offset 
    \item Docking Attempt \#1 - Failure with Safe Recovery 
    \item Pre-docking Motion to Dock Offset
    \item Docking Attempt \#2 - Success during Loss of Signal
\end{enumerate}

During the first docking attempt, the system experienced a failure. While the exact cause can not be determined from the available data, a potential localization issue based on the ISS movement is suspected (supplementary video includes entire testing footage available). Critically, the Astrobee's on-board error detection recognized that the positional error exceeded the thresholds defined for the RL policy. This triggered a pre-programmed safe recovery behavior: the execution thread automatically switched from the RL control policy back to the Astrobee's baseline controller, which maintained the robot's current pose.  This ensured a stable and safe state, preventing any uncontrolled movements or potential collisions. Following the initial failure, a second redocking attempt was initiated by a human operator, resulting in a successful dock, even during loss of signal.

\begin{figure}[h]
   \centering 
   \includegraphics[width=\linewidth]{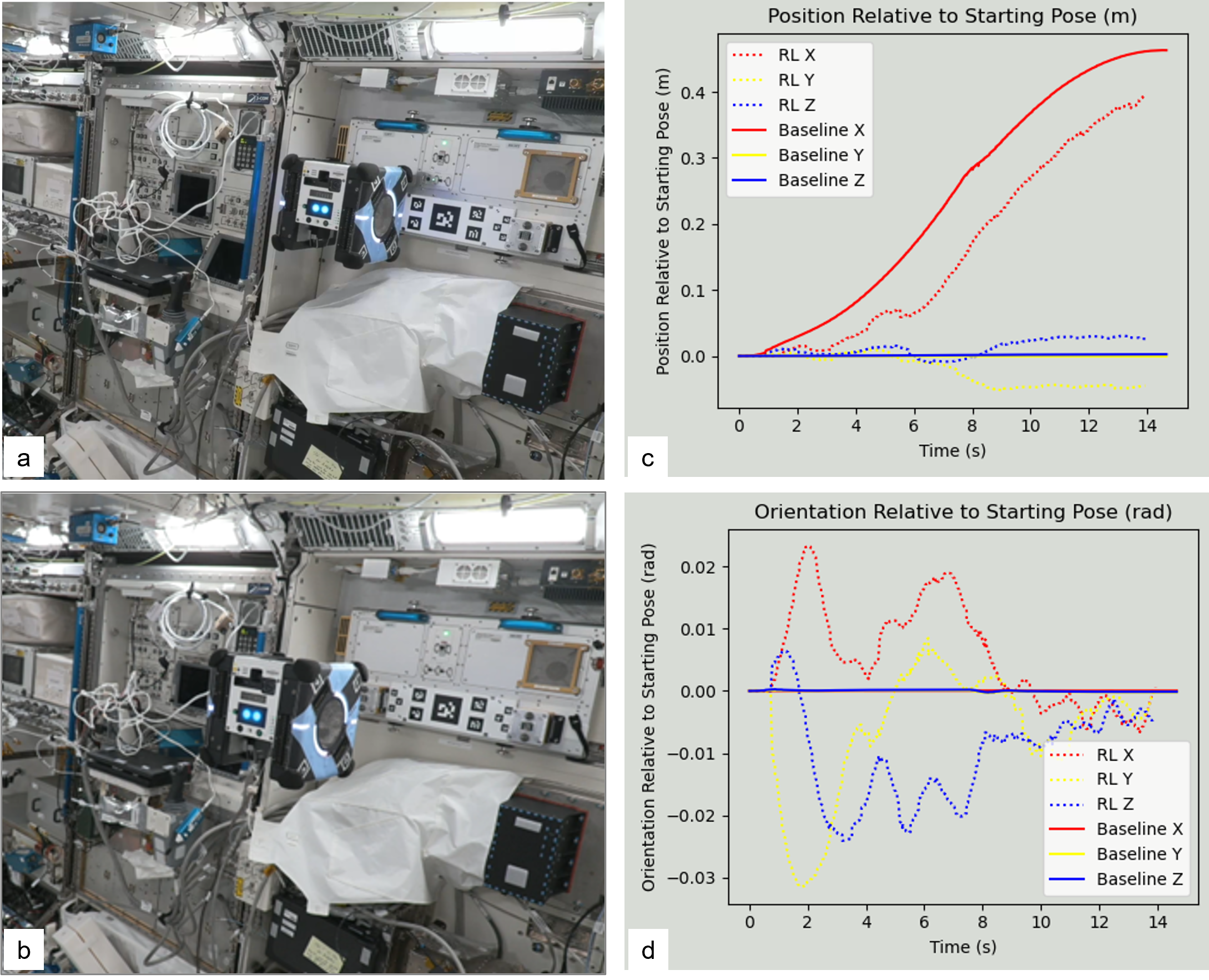}
   \caption{ISS testing of RL control on Astrobee hardware. (a) Astrobee docked starting pose. (b) Free-flying Astrobee pose after completion of undocking maneuver, 0.5 meter X-axis translation. (c) Position error from docked pose for Astrobee undock command using baseline Astrobee controller (solid lines) vs RL policy controller (dotted lines). (d) Orientation error from starting pose for Astrobee undock command.}
   \label{fig:flight} 
\end{figure}

Figure \ref{fig:flight} shows data from the undock maneuver as an example. 
The photos illustrate the starting and ending positions of the maneuver and the data shows the position and orientation error relative to the starting pose for the +0.5 meter translation in the X-axis of the RL policy (dotted line) compared to the ideal simulated Astrobee baseline controller (solid line). 
Similar to the simulated and ground testing the RL policy does not perform as well as the baseline controller but is still able to complete the desired maneuver within the Astrobee pose success criteria. 

\section{Conclusion}
\label{sec:conclusion}

This initial simulation, ground, and flight hardware testing shows the proof of concept of how a RL policy trained with NVIDIA Isaac Lab Omniverse can be used to command a free-flyer robot in 3-DOF one-G and 6-DOF zero-G environments, bridging both a sim-to-sim gap and a Sim2Real gap. 
The successful completion of simple maneuvers in a zero-G environment shows the possibility to replace classical controller with RL policies for space robotics applications. 
The APIARY testing was focused on rapidly implementing an initial example of this methodology but further refinement of the performance of the RL policy to match that of the baseline Astrobee controller could be done. 
The proposed benefit of using an RL policy instead of a classical controller is the ability to train in robustness to variation in real-life environments that cannot be replicated in a single simulation but instead require 100,000s parallel simulations to fully cover the space of possible scenarios. 

Additionally, this experiment is an example of deploying a RL policy on low size, weight and power (SWAP) hardware. The RL control had to fit within the remaining compute resources unused by the main Astrobee system.

Following this initial demonstration, more complex tasks could be undertaken in the future such as rendezvous proximity operations and docking (RPOD), grasping and manipulation using the Astrobee's 2-DOF perching arm, and adapting to unknown conditions such as changing inertial parameters when manipulating an unknown object or recovering from a thruster failure to regain motion control.

The ultimate goal of this effort is to pave the path for RL Assembly, Integration and Testing (AI\&T) practices to be used in future ISAM missions. 
In order to advance the level of autonomy used in space operations, confidence in the ability to verify performance in simulation and ground testing is paramount. 
The tasks trained and tested on this experimental Astrobee hardware will allow for early verification of basic requirements for many robotic ISAM operations, starting with the ability to control the motion of the free-flyer tested in this paper which could be used for tasks such as inspection. 
Then this methodology can be applied to expand the tasks being verified and enable future ISAM missions. 

\section{Acknowledgments}
Thanks to ONR for supporting our research. Thanks to the NASA Ames Research Center Astrobee team for helping us perform the Granite Lab testing: Ruben Garcia Ruiz, Jordan Kam, Katie Hamilton, and Roberto Carlino and thanks to the rest of the NASA Astrobee team for aid coordinating in and allowing us to do ISS testing: Jonathan Barlow, Henry Orosco, Andres Mora Vargas, Jose Benavides, Aric James Katterhagen, and Simeon Kanis. Special thanks to Kirk Hovell and the rest of the Obruta Space Solutions team for allowing us to be part of their AstroSee ISS test. Thanks to Joe Hays for setting up our lab's NVIDIA infrastructure. Thanks to NASA astronaut Anne McClain for setting up Bumble for our ISS testing!  

\bibliographystyle{IEEEtran}
\bibliography{references} 

\end{document}